\documentclass[sigconf]{acmart}
\AtBeginDocument{%
  \providecommand\BibTeX{{%
    \normalfont B\kern-0.5em{\scshape i\kern-0.25em b}\kern-0.8em\TeX}}}
\copyrightyear{2022}
\acmYear{2022}
\setcopyright{acmlicensed}\acmConference[MM '22]{Proceedings of the 30th ACM International Conference on Multimedia}{October 10--14, 2022}{Lisboa, Portugal}
\acmBooktitle{Proceedings of the 30th ACM International Conference on Multimedia (MM '22), October 10--14, 2022, Lisboa, Portugal}
\acmPrice{15.00}
\acmDOI{10.1145/3503161.3547805}
\acmISBN{978-1-4503-9203-7/22/10}

\usepackage{booktabs} 
\usepackage{tabularx}
\usepackage{multirow}
\usepackage{bm}
\usepackage{caption}
\usepackage{subcaption}
\usepackage[ruled,vlined]{algorithm2e}
\usepackage{algorithmic}
\usepackage{bbding}
\definecolor{SigmaColor}{rgb}{0.98,0.45,0.0}

\newcommand{\setParDef}{\setlength {\parskip} {0pt} }

\begin{document}

\title{Compound Batch Normalization for Long-tailed Image Classification}

\author{Lechao Cheng}
\affiliation{%
  \institution{Zhejiang Lab}
  \country{}
}
\email{chenglc@zhejianglab.com}

\author{Chaowei Fang$^\ast$}
\affiliation{%
  \institution{Xidian University}
  \country{}
}
\email{chaoweifang@outlook.com}
\thanks{$^\ast$ Corresponding author}

\author{Dingwen Zhang}
\affiliation{%
  \institution{Northwestern Polytechnical University}
  \country{}
  }
\email{zhangdingwen2006yyy@gmail.com}

\author{Guanbin Li}
\affiliation{%
  \institution{Sun Yat-sen University}
  \country{}
  }
\email{liguanbin@mail.sysu.edu.cn}

\author{Gang Huang}
\affiliation{%
  \institution{Zhejiang Lab}
  \country{}
  }
\email{huanggang@zju.edu.cn}

\renewcommand{\shortauthors}{Lechao Cheng et al.}

\begin{abstract}
Significant progress has been made in learning image classification neural networks under long-tail data distribution using robust training algorithms such as data re-sampling, re-weighting, and margin adjustment. Those methods, however, ignore the impact of data imbalance on feature normalization. The dominance of majority classes (head classes) in estimating statistics and affine parameters causes internal covariate shifts within less-frequent categories to be overlooked. To alleviate this challenge, we propose a compound batch normalization method based on a Gaussian mixture. It can model the feature space more comprehensively and reduce the dominance of head classes. In addition, a moving average-based expectation maximization (EM) algorithm is employed to estimate the statistical parameters of multiple Gaussian distributions. However, the EM algorithm is sensitive to initialization and can easily become stuck in local minima where the multiple Gaussian components continue to focus on majority classes. To tackle this issue, we developed a dual-path learning framework that employs class-aware split feature normalization to diversify the estimated Gaussian distributions, allowing the Gaussian components to fit with training samples of less-frequent classes more comprehensively. Extensive experiments on commonly used datasets demonstrated that the proposed method outperforms existing methods on long-tailed image classification. 
\end{abstract}




\maketitle
\setParDef

\setParDef
\begin{figure}
	\centering
	\includegraphics[width=0.48\linewidth]{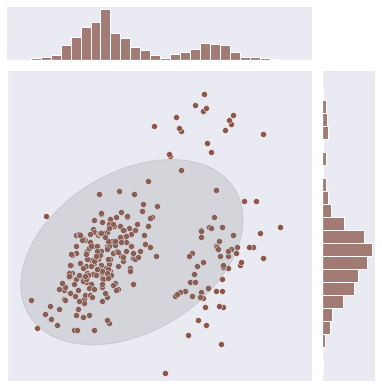}
	\includegraphics[width=0.48\linewidth]{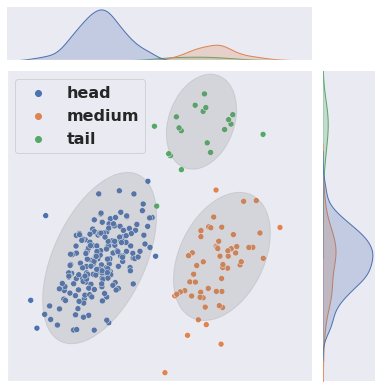}
	\caption{For an imbalanced training dataset, the conventional feature normalization (left), which leverages the single-modal Gaussian probability function  to fit the feature space, is prone to overlook samples of tail classes. Adopting multiple Gaussian distributions to fit the features (right) can mitigate the above problem.} \label{fig:teaser}
\end{figure}
\section{Introduction}
Real-world image classification data usually exhibits an imbalanced distribution due to the natural scarcity of certain classes, industry barriers, and large data collection costs. The severely imbalanced data distribution causes substantial obstruction to the learning process, considering it is difficult to balance the classification performance of head and tail classes. The imbalanced learning problem attracts extensive research interests~\cite{chawla2002smote,cui2021parametric,samuel2021distributional}. However, existing methods are incapable of deriving high accuracy on tail classes without hindering the performance of head classes or maintaining an efficient framework. This paper is targeted at learning with long-tailed training data while alleviating the above issues.

When learning deep convolutional neural networks (CNNs) with long-tailed samples, the optimization of network parameters is dominated by samples of head classes, which leads to relatively low performance for tail classes. Conventional solutions to the data imbalance problem is biasing the optimization process towards less frequent classes, such as class-balanced re-sampling~\cite{chawla2002smote,drummond2003c4}, re-weighting~\cite{huang2016learning,wang2017learning}, or classifier margin adjustment~\cite{zhong2021improving,he2021distilling}. 
However, these data rebalancing methods hamper the learning of head classes by interfering the representation capacity of CNNs. A few works attempt to address this problem through ensembling multiple classifiers learned under diverse sampling strategies~\cite{zhou2020bbn} or adopting auxiliary classifiers to highlight the learning of tail classes~\cite{xiang2020learning}. However, such methods require increased network parameters and computation burden. Besides, the impact of data imbalance on feature representation learning can not be thoroughly alleviated since they still depend on data resampling or reweighting algorithms to manage multiple classifiers. 

Batch normalization is a critical component for mitigating the internal covariate shift in the feedforward calculation process of CNNs~\cite{ioffe2015batch}. 
It can accelerate the optimization rate of network parameters and improve the generalization ability. 
Under the scenario of data imbalance, a single-modal Gaussian probability function can not fully model the feature space and is prone to overlook tail classes.
Thus the conventional batch normalization can merely eliminate the global covariate shift, but neglect the internal covariate shift of tail classes. This harms the learning efficiency and generalization capacity on tail classes.

To address the above problem, we generalize the feature normalization by modeling the feature space with compound Gaussian distributions.
As shown in Figure~\ref{fig:teaser}, the features of training samples are composed of several scattered clusters. For the purpose of fitting the features more comprehensively, we employ a compound set of mean and variance parameters to implement the feature normalization process. Every set of mean and variance parameters is applied for whitening a group of features within a local subspace, and independent affine parameters are utilized for reconstructing the distribution statistics. Such a compound feature normalization helps to eliminate the local covariate shift and alleviate the dominance of head classes. Based on the compound feature normalization, we set up the mainstream branch for the classification model and devise a moving average based expectation maximization algorithm to evaluate the statistical parameters. 

The estimation of statistical parameters in the multi-modal Gaussian probability function easily falls into local minima, where multiple Gaussian distributions still concentrate on head classes while ignoring tail classes. 
Hence, we devise a dual-path learning framework to diversify those Gaussian distributions among all classes. An auxiliary branch is set up with the split normalization, which separates classes into different subsets and processes them with independent statistical and affine parameters. This benefits to disperse statistical parameters of different Gaussian distributions. Additionally, the mainstream and auxiliary branches interact with each other via the stop-gradient based consistency constraint~\cite{chen2021exploring} for enhancing the representation learning. 
The main contributions of this paper are concluded as follows:
\begin{itemize}
\item We propose a novel compound batch normalization algorithm based on a mixture of Gaussian distributions, which can alleviate the local covariate shift and prevent the dominance of head classes.    
\item A dual-path learning framework based on the compound and split feature normalization techniques is devised to diversify the statistical parameters of different Gaussian distributions. 
\item Exhaustive experiments on commonly used datasets demonstrate significant improvement of our method compared to existing state-of-the-art methods.
\end{itemize}

\section{Related Work}
\subsection{Long-tailed Image Classification}
Real world data usually has an unbalanced distribution. Image classification models are difficult to maintain high performance on tail classes. A vast number of methods are targeted at overcoming the issue of long-tailed data distribution, which can be mainly categorized into five types including data re-sampling, data re-weighting, classifier calibration, two-stage training, and model ensembling.

\textbf{Data Re-sampling.} Oversampling tail classes~\cite{chawla2002smote} and undersampling head classes~\cite{drummond2003c4} are early methods for re-balancing the training data. \cite{huang2016learning} proposes to split samples into clusters and constructs cluster-level and class-level quintuplets to achieve re-balanced representation learning. Data augmentation by distorting images or intermediate features~\cite{wang2019implicit,chu2020feature,liu2020deep,li2021metasaug} can be utilized for expanding the  sample sizes of tail classes. However, these methods easily lead to under-fitting of head classes or over-fitting of tail classes.

\textbf{Data Re-weighting.} The other type of commonly used methods is increasing weighting coefficients when calculating training losses for samples of tail classes. 
Simple data re-weighting can be implemented with the inverse class frequencies~\cite{huang2016learning,wang2017learning}. \cite{cui2019class} devises a more reasonable way to estimate the effective number of samples and incorporate it into the cross entropy loss.
Focal loss~\cite{lin2017focal} is capable of concentrating on `hard' samples and can also benefit the learning of tail classes. 
\cite{park2021influence} re-weights individual samples with influence factors estimated from gradients of network parameters.
\cite{cui2021parametric} devises a novel contrastive loss based on center learning and attempts to incorporate it with the balanced soft-max function for addressing the data imbalance issue.

\textbf{Classifier Calibration.}
Another type of data imbalance learning methods focus on calibrating the supervision signals, decision margins, or parameters of classifiers .
\cite{zhong2021improving} smooths the one-hot label vectors and relieves the over-confidence on head classes.
\cite{he2021distilling} leverages predictions of the teacher model to rectify the label distribution.
\cite{chen2021supercharging} transfers the knowledge of head classes to tail classes considering the invariant label-conditional features across different labels.
\cite{cao2019learning} shifts the decision boundary to head classes, thus improving the generalization error for tail classes without influencing the performance of head classes.
\cite{samuel2021distributional} devises a distributionally robust loss by penalizing distances between samples and empirical category centroids.
\cite{zhang2021distribution} utilizes an adaptive module to directly adjust the classification scores.
\cite{hong2021disentangling} compensates the prediction logits for alleviating the label distribution shift between source and target data. 
\cite{liu2021gistnet} utilizes a set of classifier displacement vectors to transfer the geometry of head classes to tail classes.
\cite{kini2021label} devises a vector-scaling loss to unify the advantages of additive and multiplicative logit adjustments.
\cite{xu2021towards} resorts to the mixup algorithm to encourage the occurrence of sample pairs from head and tail classes, and compensates the cross entropy with the Bayes bias.

\textbf{Multi-Stage Training.}
Deferring the re-balancing procedure helps to relieve the intrinsic artifacts of data re-balancing methods, such as over-fitting with minority classes caused by data re-sampling and optimization instability caused by loss reweighting~\cite{cao2019learning}. \cite{zhou2020bbn} constructs a two-branch framework to combine the instance-balanced learning and class-balanced learning in the cumulative manner. 
\cite{li2021self} employs three cascaded training stages, including self-supervised feature learning, class-balanced learning, and instance-balanced learning under the guidance of the knowledge distilled from the second stage.  

\textbf{Model Ensembling.}
A few imbalance learning algorithms aim at combining the advantages of multiple models separately trained with different subsets of samples.  
\cite{xiang2020learning} splits classes into a few subsets, learns an expert model for each class subset, and integrates different expert models to teach the student model.
\cite{cai2021ace} further devises a distribution-aware class splitting planner to increase the exposure of tail classes among expert models.
\cite{wang2020long} trains multiple diversified expert models simultaneously and sets up a routing mechanism to prune the multi-expert system for reducing the computation cost.
\cite{zhang2021test} builds up multiple expert classifiers guided with conventional or balanced loss functions. It also attempts to leverage the cross-augmentation prediction consistency to improve the generalization of learned expert models on testing data with unknown distributions. 
The main drawback of this kind of methods is that learning multiple models inevitably increases the computation burden during training or testing.

\subsection{Feature Normalization}
Feature normalization, such as batch normalization~\cite{ioffe2015batch}, layer normalization~\cite{ba2016layer}, and group normalization~\cite{wu2018group}, is commonly applied for eliminating the covariate shift in various tasks~\cite{cheng2018intrinsic,zhang2020weakly,cheng2021boundary,zhang2021weakly,zhang2021token,hao2022group}. Such kind of operations benefit to accelerate the optimization process, prevent over-fitting, and relieve the gradient vanishing/explosion phenomenon. However, these methods utilize single Gaussian distribution, namely one set of mean and variance, to model the input features. 
The practical feature points usually exhibit a multi-modal Gaussian distribution. 
When the training samples are severely imbalanced, fitting them with a single-modal distribution leads to overlook of tail classes, which harms the efficacy of the normalization in learning tail classes. To address this issue, we design a novel feature normalization method based on the multi-modal Gaussian distribution. The momentum-based expectation maximization algorithm is incorporated for estimating multiple means and variances. Similar to our approach, SL-BN~\cite{zhong2021improving} proposes to update the mean and variance variables of batch normalization layers in the deferred training stage with class-balanced resampling. Our proposed compound batch normalization (CBN) differs from SL-BN in that, Gaussian mixtures are used to model the feature space in CBN while SL-BN still relies on single Gaussian distribution. Targeted at preventing pure noise images from distorting the estimation of the feature distribution, DAR-BN~\cite{zada2021pure} splits the mean and variance variables for normal images and pure noise images. Auxiliary BN~\cite{merchant2020does} is used for alleviating the disparity between training images processed with strong augmentation and testing images by setting up an extra batch normalization branch for strongly augmented images. The target of our devised CBN is distinct to them. CBN aims at modeling training data with multiple Gaussian distributions and preventing overlooking samples of tail classes during the batch normalization. Besides, the core algorithmic principle of our CBN is different to that of DAR-BN and Auxiliary BN. DAR-BN and Auxiliary BN essentially depends on a main branch to implement the feature normalization for input images. CBN accomplishes the normalization process by adaptively accumulating the normalization results calculated with individual Gaussian components.


\section{Approach}

\subsection{Preliminary Knowledge on Batch Normalization}
First, we remind the calculation procedure of batch normalization. Suppose the input feature map be $\mathbf X\in \mathbb R^{B\times D\times H\times W}$, where $B$, $D$, $H$, and $W$ denote the batch size, number of channels, height, and width, respectively. We can flatten dimensions except for channels into a single dimension, resulting in a two-dimensional tensor $\mathbf x\in \mathbb R^{D \times N}$ ($N=BHW$). Channel-wise statistical variables including mean $\hat{\bm\mu}\in\mathbb R^{D\times 1}$ and variance $\hat{\bm\sigma}^2\in \mathbb R^{D\times1}$ are estimated as,
\begin{align}
    \hat{\bm\mu}&=\frac{1}{N} \sum_{i=1}^N \mathbf x_i, \\
    \hat{\bm\sigma}^2&= \frac{1}{N} \sum_{i=1}^N (\mathbf x_i-\hat{\bm\mu})^2,
\end{align}
where $\mathbf x_i\in\mathbb R^{D\times1}$ represents the feature of the $i$-th point. During the network optimization, these statistical variables are accumulated with the moving average operation:
\begin{align}
    \bm{\mu} &:= \lambda \bm\mu + (1-\lambda) \hat{\bm\mu}, \\
    {\bm\sigma}^2 &:= \lambda {\bm\sigma}^2 + (1-\lambda) \hat{\bm\sigma}^2.
\end{align}
$\lambda$ is the momentum factor, determining the updating rate of $\bm\mu$ and $\bm\sigma$.  Afterwards, the input feature $\mathbf x$ is normalized as below:
\begin{equation}
    \hat{\mathbf x} = \bm\Sigma^{-\frac{1}{2}}(\mathbf x-\bm\mu),
\end{equation}
where $\epsilon$ denotes a constant. $\bm\Sigma= \textrm{diag}(\bm \sigma^2)$, where $\textrm{diag}(\cdot)$ transforms the input vector into a diagonal matrix. Finally, the feature values are scaled and shifted with affine parameters $\gamma$ and $\beta$,
\begin{equation}
    \mathbf x^\prime= \gamma \hat{\mathbf x}+\beta.
\end{equation}
The batch normalization operation can remove the internal covariate shift during the feedforward propagation of neural networks, which benefits to stabilizing and accelerating the training speed. However, when the training samples are severely unbalanced, the calculation of the statistical variables is dominated by head classes, and the learning of affine parameters is also biased towards them. This induces to overlook of tail classes in the normalization process, which hinders the learning on less frequent classes.

\subsection{Compound Batch Normalization}\label{sec:cn}

For modeling the feature space more comprehensively, we adopt a mixture of statistical variables to fit the feature distribution. Suppose the number of Gaussian distributions be $M$. The $j$-th Gaussian distribution is represented with a triplet of variables, including prior probability,  mean, and variance variables, which are defined by $\tau_j$, $\bm\mu_j$, and $\bm\Sigma_j$, respectively. $\bm\Sigma_j\in\mathbb R^{D\times D}$ is the diagonal variance matrix. Then, the batch normalization can be extended to compound probability distributions. The input feature is normalized according to $M$ Gaussian distributions independently, deriving of $M$ normalization branches. Provided a feature vector $\mathbf x_i$, the normalization with the $j$-th Gaussian distribution is formulated as follows,
\begin{equation}
    \hat{\textbf{x}}^{(j)}_i=\bm\Sigma^{-\frac{1}{2}}_j(\mathbf x_i-\bm\mu_j).
\end{equation}

For the estimation of compound statistic variables,  we apply the moving average to implement the expectation-maximization algorithm. First, we calculate the probability values of every feature vector in  $M$ Gaussian distributions. Then, a temporary set of prior probability, mean, and variance variables is estimated based on input features and their probability values for the current batch of samples. These temporary variables are accumulated in the moving average manner. The algorithmic detail of the estimation process is introduced below.

\noindent \textbf{Expectation Step.} Based on the previously estimated statistic variables, we can calculate the probability values of all input feature vectors with respect to $M$ Gaussian distributions.
The probability of $\mathbf x_i$ belonging to the $j$-th distribution is estimated as follows,
\begin{equation}
    w_{ij}= \frac{ \tau_j f(\mathbf x_i,\bm\mu_j,\bm\Sigma_j) }{\sum_{k=1}^M \tau_k f(\mathbf x_i,\mu_k,\Sigma_k) }.
\end{equation}
$f(\cdot)$ is the Gaussian probability density function, which is formulated as follows,
\begin{equation}
    f(\mathbf x_i,\bm\mu_j,\bm\Sigma_j)=\frac{\exp(-\frac{1}{2}(\mathbf x_i-\bm \mu_j)^T\bm\Sigma^{-1}_j (\mathbf x_i-\bm \mu_j))}{\sqrt{(2\pi)^M|\bm\Sigma_j|}}.
\end{equation}
\begin{figure}[t]
\centering
\includegraphics[width=\linewidth]{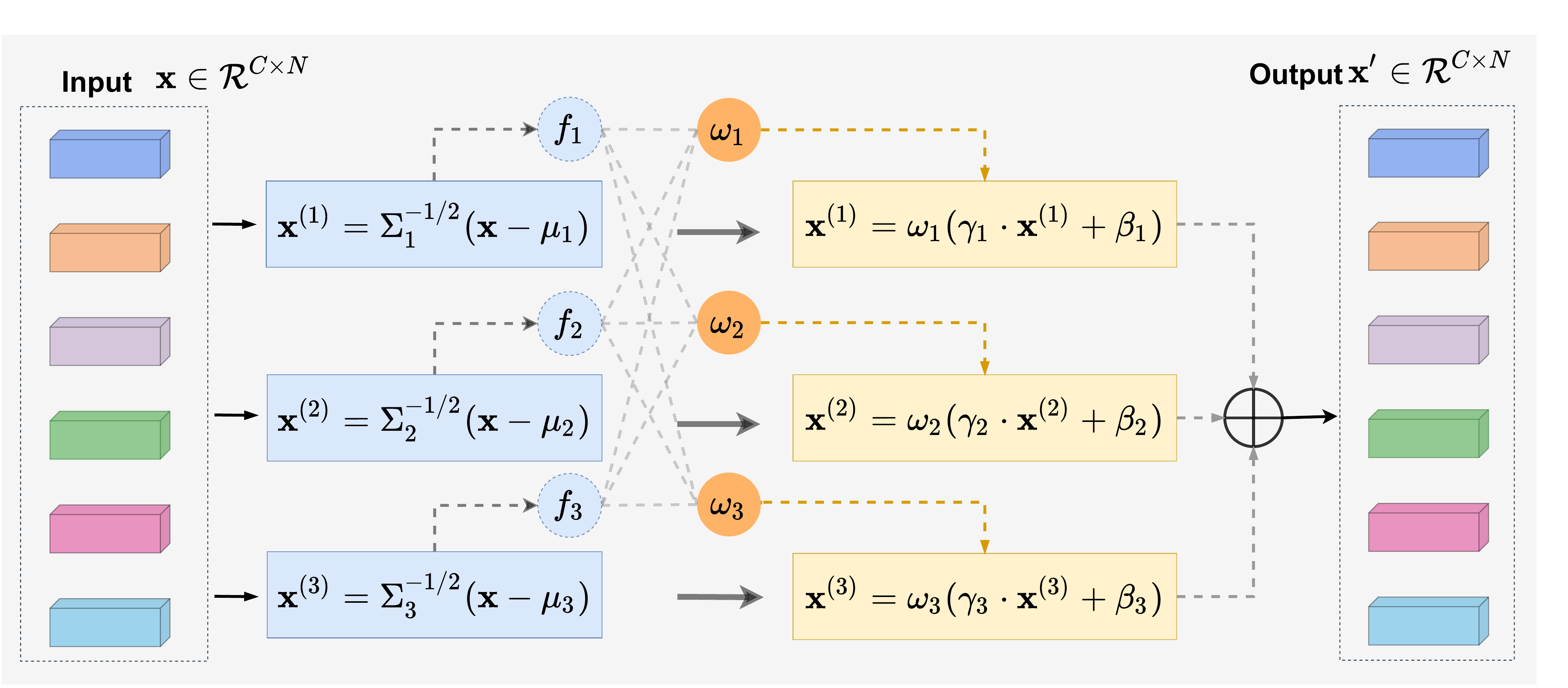}
\caption{Compound batch normalization based on a mixture of Gaussian distributions.}\label{fig:cn}
\vspace{-2mm}
\end{figure}

\begin{figure}[t]
\centering
\includegraphics[width=\linewidth]{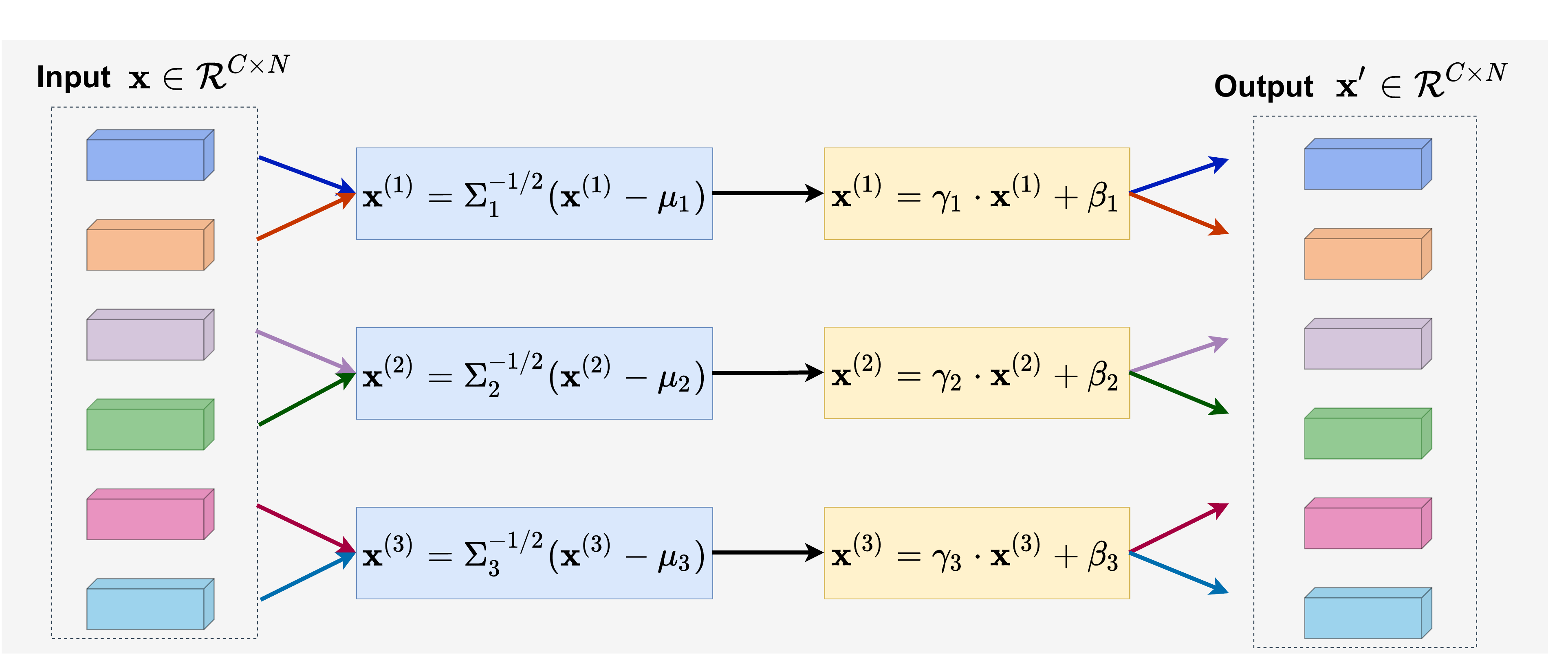}
\caption{Split batch normalization.}\label{fig:sn}
\vspace{-2mm}
\end{figure}


\noindent \textbf{Maximization Step.} Temporary statistic parameters are estimated according to the probability values for each training batch. Practically, the temporary prior probability $\hat{\tau}_j^{c}$, mean $\hat{\bm \mu}_j^c$, and variance $\hat{\bm \sigma}_j^c$ of the $j$-th Gaussian distribution are calculated as below,
\begin{align}
    \hat{\tau}_j^c &= \frac{1}{N} \sum_{i=1}^N w_{ij}, \\
    \hat{\bm \mu}_j^c &= \frac{\sum_{i=1}^N w_{ij} \mathbf x_i}{\sum_{i=1}^N w_{ij}},\\
    \hat{\bm \sigma}_j^c &= \frac{\sum_{i=1}^N w_{ij} (\mathbf x_i-\hat{\bm \mu}_j)^2}{\sum_{i=1}^N w_{ij}}.
\end{align}

\noindent \textbf{Temporal Accumulation.} Similar to the conventional batch normalization, the moving average is applied to accumulate the above temporary variables, namely, 
\begin{align}
\tau_j      &:=\lambda^c \tau_j + (1-\lambda^c) \hat{\tau}_j^c, \\
\bm\mu_j    &:=\lambda^c\bm\mu_j+(1-\lambda^c)\hat{\bm \mu}_j^c, \\
\bm\Sigma_j &:=\lambda^c \bm\Sigma_j+(1-\lambda^c)\textrm{diag}(\hat{\bm \sigma}_j).
\end{align}
Here, $\textrm{diag}(\cdot)$ transforms the input vector into a diagonal matrix, and $\lambda^c$ is a constant.

Finally, separate scaling  and bias coefficients are learned to redistribute the values of different normalization branches. Those redistributed values are combined by the following weighted summation operation,
\begin{equation}
    \mathbf x^\prime_i = \sum_{j=1}^M w_{ij}(\gamma_j \hat{\textbf{x}}^{(j)}_i + \beta_j).
\end{equation}
$\gamma_j$ and $\beta_j$ represent the scaling and bias coefficient of the $j$-th normalization branch respectively.
The calculation process of the above normalization algorithm is illustrated in Figure~\ref{fig:cn}. More details can be found in Appendix~\ref{sm:algo}.

\begin{figure*}[t]
\centering
\includegraphics[width=0.9\linewidth]{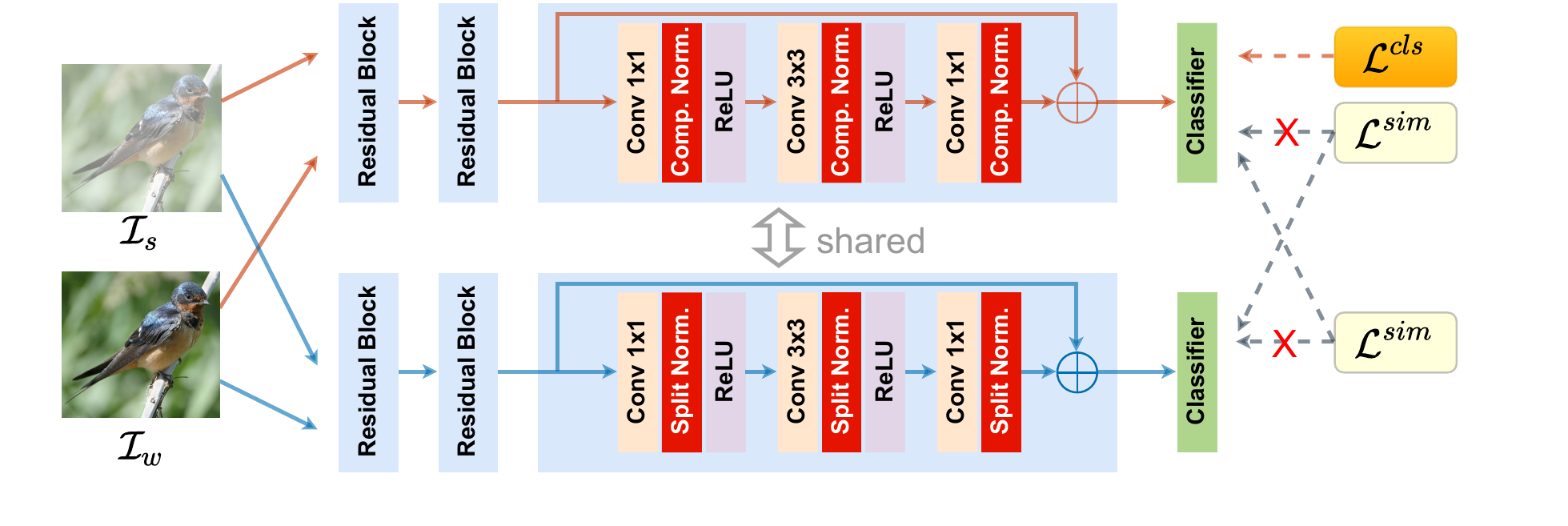}
\vspace{-1mm}
\caption{The proposed dual-path learning framework.}\label{fig:framework}
\end{figure*}

\subsection{Split Batch Normalization}
The generalized batch normalization can be easily incorporated into existing convolutional neural networks by replacing their original normalization layers. 
However, learning compound distributions with the expectation-maximization algorithm may easily fall into local optima and suffer from model collapse. 
To overcome this problem, we set up a split normalization strategy to diversify the Gaussian distributions with different sets of training samples.

First, we uniformly split all class labels into $M$ independent groups according to their serial numbers, resulting in $\{\mathbb C^j\}_{j=1}^M$, where $\mathbb C^j$ represents the $j$-th set of class labels. For $j\neq k$, $\mathbb C^j \cap \mathbb C^k=\varnothing$. The union of all class sets ($\cup_{j=1}^M \mathbb C^j$) is equal to the sequence of integers from 1 to the number of classes ($K$).
According to these class sets, the input features in $\mathbf x$ can be separated into $M$ groups as well, namely $\{ \mathbf x ^{j} \in \mathbf R^{D\times N_j}\}_{j=1}^M$, where the features in $\mathbf x ^{j}$ come from images with class labels in $\mathbb C^j$, and $N_j$ represents the number of features in the $j$-th group.
Then, a split normalization strategy illustrated in Figure~\ref{fig:sn} is devised to process $M$ sets of features with $M$ Gaussian distributions, respectively. Meanwhile, each set of features is utilized for calculating the other temporary mean and variance,
\begin{align}
    \hat{\bm \mu}_j^s &= \frac{1}{n^j} \sum_{i=1}^{n^j} \mathbf x_i^j,\\
    \hat{\bm \sigma}_j^s &= \frac{1}{n^j} \sum_{i=1}^{n^j} (\mathbf x_i^j-\hat{\bm \mu}_j^s)^2.
\end{align}
The above variables are also leveraged to update $\bm \mu_j$ and $\bm \Sigma_j$: $\bm \mu_j:=\lambda^s \bm \mu_j + (1-\lambda^s) \hat{\bm \mu}_j^s $, and $\bm \Sigma_j:=\lambda^s \bm \Sigma_j + (1-\lambda^s) \textrm{diag}(\hat{\bm \sigma}_j^s)$. This process is beneficial for diversifying multiple Gaussian distributions and preventing the distribution collapse issue. The calculation process of the split batch normalization is illustrated in Algorithm~\ref{alg:sbn} (Appendix~\ref{sm:algo}).

\subsection{Dual-Path Learning}\label{sec:dp}
As shown in Figure~\ref{fig:framework}, we build up the classification model with ResNet~\cite{he2016deep}, which is learned with two branches.
In the top branch, the compound normalization is applied for implementing the network feedforward process, while the split normalization is adopted for standardizing intermediate features in the bottom branch.
Given an input image $\mathbf I\in\mathbb R^{3\times H\times W}$, we utilize a series of weak augmentation operations to transform it into $\mathbf I_{weak}$, and the other series of strong augmentation operations is leveraged to produce the other variant of the input image denoted by $\mathbf I_{strg}$. Feeding $\mathbf I_{weak}$ and $\mathbf I_{strg}$ into the branch based on the compound normalization, we can obtain predicted class-wise logits $\mathbf o^c_{weak}\in \mathbb R^{K\times1}$ and $\mathbf o^c_{strg}\in \mathbb R^{K\times1}$, respectively. $K$ denotes the number of classes.
The split normalization branch also derives class-wise logits $\mathbf o^s_{weak}$ and $\mathbf o^s_{strg}$ from $\mathbf I_{weak}$ and $\mathbf I_{strg}$, respectively. The maximization of the similarity between predictions of two branches is employed for optimizing network parameters,
\begin{equation} \label{eq:loss-cos}
    L^{sim} = -\mathcal S(\mathbf o^c_{strg}, \textrm{detach}(\mathbf o^s_{weak}))-\mathcal S(\mathbf o^s_{strg}, \textrm{detach}(\mathbf o^c_{weak})).
\end{equation}
\begin{algorithm}[t]
\caption{Algorithm for one training epoch of the dual learning framework.} \label{alg:dual}
\SetAlgoLined
\SetKwInput{Inputs}{Input}
\Inputs{Training images: $\{\mathbf I_l\}_{l=1}^L$, and their labels: $\{\mathbf y_l \}_{l=1}^L$; $M$ sets of class labels: $\{\mathbb C^j\}_{j=1}^M$; $M$ sets of Gaussian statistic variables: $\{\tau_j\}_{j=1}^M$, $\{\bm\mu_j\}_{j=1}^M$, and $\{\bm\Sigma_j\}_{j=1}^M$. }
\SetKwInput{Output}{Output}
\Output{Optimized network parameters.}
\begin{algorithmic}[1]
\STATE Shuffle training images and separate into minibatches with size of $B$;
\FOR{$i=1$ \KwTo $\frac{L}{B}$}
    \STATE Fetch a batch of training images  $\mathbf B \in \mathbb R^{B\times3\times H \times W}$;
    \STATE Distort images in $\mathbf B$ with weak augmentation operations, resulting in $\mathbf B_{weak}$;
    \STATE Feed $\mathbf B_{weak}$ through the compound and split network paths, resulting in $\mathbf O^c_{weak}$ and $\mathbf O^s_{weak}$, respectively;
    \STATE Distort $\mathbf B$ with strong augmentation operations, deriving of $\mathbf B_{strg}$;
    \STATE Feed $\mathbf B_{strg}$ through the compound and split network paths, resulting in $\mathbf O^c_{strg}$ and $\mathbf O^s_{strg}$, respectively;
    \STATE Calculate training losses according to Eq. (\ref{eq:loss-cos}) and (\ref{eq:loss-cls});
    \STATE Update network parameters by stochastic gradient descent;
\ENDFOR
\end{algorithmic}
\vspace{-0.5mm}
\end{algorithm}
The similarity metric $\mathcal S(\mathbf o_1, \mathbf o_2)$ is implemented with the cosine function, $\mathcal S(\mathbf o_1, \mathbf o_2)=\mathbf o_1^T \mathbf o_2/(||\mathbf o_1||_2||\mathbf o_2||_2)$. `$\textrm{detach}(\mathbf o)$' represents the stop gradient operation, which means $\mathbf o^s_w$ and $\mathbf o^c_w$ are regarded as constants. 

Finally, the balanced softmax function is employed for calculating the training loss on the network prediction of strongly augmented image $\mathbf I_{strg}$,
\begin{equation} \label{eq:loss-cls}
    L^{cls} =  -\log(\frac{ n_y \exp(o^c_{strg}[y]) }{ \sum_{i=1}^K n_i \exp(o^c_{strg}[i])} ).
\end{equation}
$n_i$ denotes the number of samples in the $i$-th class, and $o^c_{strg}[i]$ is the $i$-th element in $\mathbf o^c_{strg}$. $y$ represents the ground-truth label of the input image $\mathbf I$.  The practical training procedure is implemented with the minibatch size of $B$. One training epoch of the dual learning framework is summarized in Algorithm~\ref{alg:dual}.

In training stage, the running variables(mean,variance and prior probability) of each Gaussian distribution are updated with temporal variables. During testing, the running variables are fixed, and only the calculation path of compound batch normalization is preserved while the path of split batch normalization is no longer used in the inference process of the model.

\section{Experiments}
We test the effectiveness of the proposed strategy on representative synthetic data as well as real-world datasets in this section. Table~\ref{tab:data} describes the details of long-tailed data used in this work. The evaluation metric for image classification is top-1 accuracy (\%).

\subsection{Datasets}

\begin{itemize}
    \item \textbf{CIFAR10-LT/100-LT.} 
     CIFAR10-LT/100-LT is formed by resampling images from the original CIFAR10-LT/100~\cite{krizhevsky2009learning} dataset. 
     The class-wise sample sizes obey an exponential distribution.
     We denote the imbalance ratio as $\rho = N_{max}/N_{min}$, where $N_{max}$ and  $N_{min}$ is the sample size of the most frequent class and the least frequent class, respectively. We validate the performance of all models under three settings for $\rho$ ($\in \{100,50,10\}$).
    \item \textbf{ImageNet-LT.} ImageNet-LT is a subset of the ImageNet1K~\cite{russakovsky2015imagenet} dataset that contains images from 1000 categories, with a maximum of 1280 images per class and a minimum of 5 images per class. The dataset consists of 115.8k training images, 20k validation images, and 50k test images.
    
    \item \textbf{Places-LT.} Places-LT features an unbalanced training set from Places-2~\cite{zhou2017places}, with 62,500 images for 365 classes. The class frequencies are distributed according to a natural power law with a maximum of 4,980 images per class and a minimum of 5. The validation and testing sets are evenly distributed, with 20 and 100 images per class in each.
    
    \item \textbf{iNaturalist2018} There are 437K images in iNaturalist-2018~\cite{van2018inaturalist}, with 6 degrees of label granularity. This dataset is challenging since the labels are long-tailed and fine-grained. We only evaluate the most granular descriptors (species), resulting in 8142 distinct classes with a naturally imbalanced distribution.
\begin{table}[htp]
\centering

\caption{Introduction of long-tailed datasets used in the experiments. Noted that $*$ means original data size in CIFAR10 and CIFAR100}\label{tab:data}
\begin{tabular}{l | c | c| c|c | c} 
\toprule
Dataset & Classs & $\rho$ & Train & Val. & Test \\ \midrule
CIFAR10-LT &  10 & 10-100 & 50k$^{*}$ & - & 10k \\
CIFAR100-LT &  100 & 10-100 & 50k$^{*}$ & - & 10k \\
ImageNet-LT &  1000 & 256 & $\sim$115.8k & 20k & 50k \\
Places-LT &  365 & 996 & 62.5k & 7.3k & 36.5k \\
iNaturalist2018 &  8142 & 500 & $\sim$437.5k & $\sim$24.4k & $\sim$149.4k \\
\bottomrule
\end{tabular}
\vspace{-4mm}
\end{table}   
\end{itemize}
\subsection{Implementation Detail}
Our method is implemented with PyTorch~\cite{paszke2019pytorch}. The weak augmentation is composed of random cropping and flipping, whereas AutoAugment~\cite{cubuk2019autoaugment} is used to generate strongly augmented images. We use SGD as the optimizer, with a learning rate of 0.05 that decays concerning the cosine annealing schedule. The number of training epochs is set to 400, and the mini-batch size is set to 128. $\lambda$, $\lambda^c$ and $\lambda^s$ are all set to 0.1. Without specification, the backbone is ResNet32, and all models are trained from scratch by default.

\subsection{Ablation Studies}
This section examines the efficacy of each component of the proposed approach and provides a detailed experimental study.\\

\begin{table*}[t]
\caption{Analysis of different components. The experiments are conducted on CIFAR10-LT and CIFAR100-LT, and we train the model from scratch for all cases. The `Baseline' models are trained with AutoAugment~\cite{cubuk2019autoaugment} transforms and balanced loss~\cite{Ren2020balms}. `SBN' and `CBN' indicate split batch normalization and compound batch normalization respectively. `DPL' means dual path learning mentioned in Section~\ref{sec:dp}. The number of mixtures M is empirically set to 4 for \textbf{CBN}.}
\resizebox{0.9\textwidth}{!}{
\centering
\label{tab:ablation}
\setlength\tabcolsep{2.pt}
\begin{tabular}  {l|c|c|c|c|c|c}
\toprule
\multirow{2}{*}{Method} & \multicolumn{3}{c|}{CIFAR10-LT} & \multicolumn{3}{c}{CIFAR100-LT}\\
\cmidrule(l){2-4} \cmidrule(l){5-7}  

\multirow{2}{*}{} & \multicolumn{1}{c|}{ $\rho$=100} & \multicolumn{1}{c|}{ $\rho$=50} & \multicolumn{1}{c|}{ $\rho$=10} & \multicolumn{1}{c|}{ $\rho$=100} & \multicolumn{1}{c|}{ $\rho$=50} & \multicolumn{1}{c}{ $\rho$=10}  \\ 



\midrule


\textbf{Baseline}  & 81.87 & 84.65  & 88.34 & 49.66 & 52.18 & 62.49  \\

\textbf{Baseline} + \textbf{SBN} & 82.03\textcolor{red}{(+0.16)}   & 84.73\textcolor{red}{(+0.08)} &  88.89\textcolor{red}{(+0.55)} & 49.86\textcolor{red}{(+0.20)}  & 52.44\textcolor{red}{(+0.26)}  & 62.55\textcolor{red}{(+0.06)}    \\

\textbf{Baseline} + \textbf{CBN} & 84.12\textcolor{red}{(+2.25)} & 86.75\textcolor{red}{(+2.10)} &  89.89\textcolor{red}{(+1.55)}& 52.16\textcolor{red}{(+2.50)} & 57.26\textcolor{red}{(+5.08)} & 64.37\textcolor{red}{(+1.88)} \\

\textbf{Baseline} + \textbf{CBN} + \textbf{SBN} &  84.31\textcolor{red}{(+2.44)}  & 87.21\textcolor{red}{(+2.56)}& 90.44\textcolor{red}{(+2.10)}  & 52.76\textcolor{red}{(+3.10)} & 58.04\textcolor{red}{(+5.86)} & 64.97\textcolor{red}{(+2.48)} \\ 

\textbf{Baseline} + \textbf{CBN} + \textbf{SBN} + \textbf{DPL} & 84.98\textcolor{red}{(+3.11)}  &88.70\textcolor{red}{(+4.05)} & 91.82\textcolor{red}{(+3.48)} & 53.31\textcolor{red}{(+3.65)} & 58.13\textcolor{red}{(+5.95)} & 65.35\textcolor{red}{(+2.86)}  \\ \bottomrule
\end{tabular}
}
\end{table*}

\noindent\textbf{Components Analysis}
We conduct comprehensive ablation research to validate the essential components of our framework. Table~\ref{tab:ablation} shows the results of the experiments. The loss criteria for evaluating the consistency between predictions and provided labels is the Balanced Softmax Cross-Entropy~\cite{Ren2020balms}. As can be seen in Table~\ref{tab:ablation}, using \textbf{CBN} (compound batch normalization) improves performance significantly. For example, on the CIFAR10-LT dataset with $\rho=100$, the \textbf{CBN} results in a 2.25\% increase in accuracy. The \textbf{SBN} (split batch normalization) assists in diversifying the statistical variables of multiple Gaussian distributions. The \textbf{DPL} (dual-path learning) approach, which was devised for feature learning, results in considerable performance gains. When $\rho$ is set to 100, 50, and 10, the accuracy of CIFAR100-LT is raised by 0.67\%, 1.49\%, and 1.38\%, respectively. 

\begin{figure}[t]
	\centering
	\includegraphics[width=\linewidth]{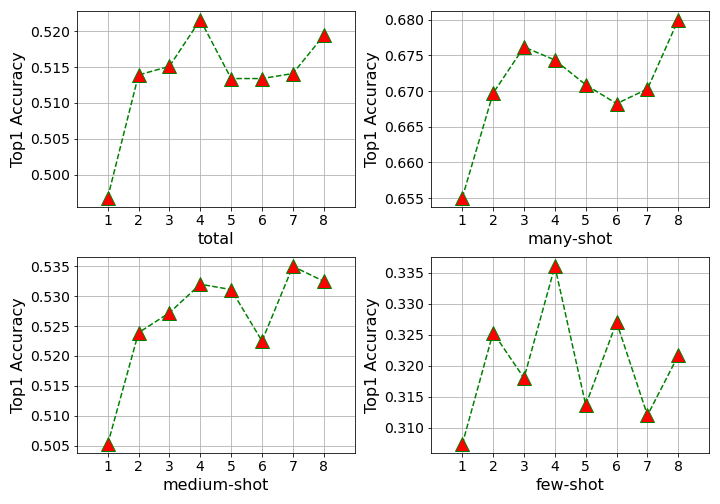}
	\caption{Performance of models trained with various numbers of Gaussian distributions on three disjoint subgroups. CIFAR100-LT with $\rho=100$ is used for training and testing. 
	}
	\label{fig:num_mix}
	\vspace{-5mm}
\end{figure}

\noindent\textbf{Single-Modal Gaussian vs.\ Multi-Modal Gaussian.}
We first highlight the significance of compound normalization mentioned in section~\ref{sec:cn}. We follow~\cite{Ren2020balms} to evaluate the classification accuracy on three disjoint sets of classes: many-shot (classes with more than 100 training samples), medium-shot (classes with 20–100 training samples), and few-shot (classes with fewer than 20 training samples). Figure~\ref{fig:num_mix} shows the top-1 accuracy against the number of mixtures on CIFAR100-LT. Here, balanced loss and dual-path learning are employed in this comparison. The plot demonstrates that estimating multiple Gaussian distributions favors all three subgroups. As can be seen that, multiple Gaussians (M>1) are superior to single Gaussian (M=1) while too large M (larger than 4) cannot bring continuous performance gain since the difficulty of parameter estimation increases as M grows up. Apart from this, our proposed method causes subtle increase of computational burden compared to the baseline method. For example, when ResNet32 is used as the backbone, our model (M=4) occupies the space of 0.464M compared to original 0.461M in memory. Besides, the training process of the baseline and our approach consumes ~3 hours and ~4 hours respectively on CAFIR10-LT under imbalance factor of 100. The inference time difference between our method and the baseline can be very small.

\begin{figure}[t]
    \vspace{-2mm}
	\centering
	\includegraphics[width=\linewidth]{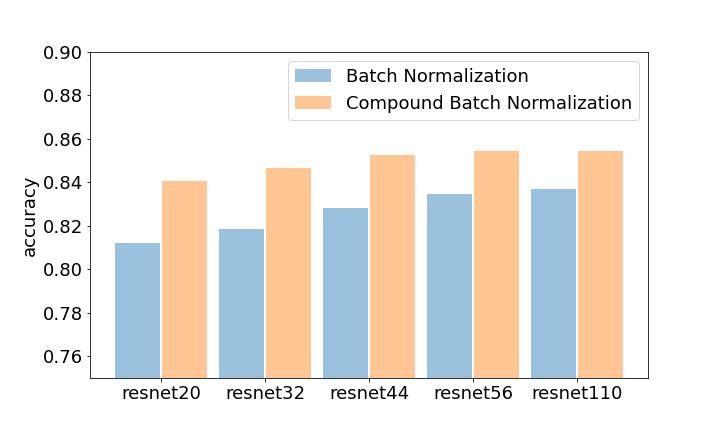}
	\caption{We test our technique on CIFAR10-LT with an imbalance factor of 100 and the number of mixtures is set to 4.}
	\label{fig:backbone}
	\vspace{-3mm}
\end{figure}

\noindent\textbf{Different Backbones.}
To test whether our method can generalize to other network architectures, we try to apply it to various variants of ResNet. The experimental results are presented in Figure~\ref{fig:backbone}.
Unlike conventional residual networks, ResNet-20/32/44/56/110 is built upon three hyper-blocks. 
As can be observed, the proposed compound batch normalization consistently outperforms the conventional batch normalization across network architectures.

\noindent\textbf{Combination with Re-sampling and Re-weighting}
We study the impact of CBN on data re-sampling/re-weighting algorithms in this subsection, by training our devised model with those algorithms. We accomplish decoupling training by decoupling the learning procedure into representation learning (80\% epochs) and classification learning (20\% epochs). The re-sampling/re-weighting method is only applied for fine-tuning parameters of classifiers during the classification learning stage while the parameters of the feature extractor are fixed. The backbone used in Table~\ref{tab:comp-rsrw} is ResNet32, and we evaluate the top-1 accuracy on three disjoint subgroups on CIFAR100 with an imbalance factor of 100. We set (I) as the baseline that the model is trained with AutoAugment~\cite{cubuk2019autoaugment} transforms and balanced loss~\cite{Ren2020balms}. Experiment (II) demonstrates that merely using  re-sampling and decoupling training is ineffective in improving the accuracy. On the other hand, experiments (III) and (V) boost the baseline performance by promoting the accuracy on tail classes, but degrade the peformance on head classes. For our compound normalization, it delivers significant increases for most classes (except for 'Many' in (VIII) and 'Few' in (VI)) when combining re-weighting or re-sampling with decoupling training.

\begin{table}
\centering
\caption{Performance of our compound batch normalization (\textbf{CBN}) incorporated with data re-sampling (\textbf{RS}), data re-weighting (\textbf{RW}), and decoupling training (\textbf{DT}). }\label{tab:comp-rsrw}

\setlength\tabcolsep{3.8pt}
\begin{tabular}{ c c c cccc cc } 
\toprule

\textbf{EXP}  & \textbf{CBN} & \textbf{DT} & \textbf{RW} & \textbf{RS} & \textbf{Total} & \textbf{Many} & \textbf{Medium} & \textbf{Few}\\ 
\midrule
(I) & \XSolidBrush & \XSolidBrush & \XSolidBrush & \XSolidBrush & 49.66 & 66.93 & 51.20 & 28.24\\
(II) & \XSolidBrush & \CheckmarkBold & \XSolidBrush & \CheckmarkBold & 49.62 & 67.17 & 50.61 & 28.54\\
(III) & \XSolidBrush & \CheckmarkBold & \CheckmarkBold & \XSolidBrush &  50.56 & 62.49 & 52.08 & 35.21\\
(IV) & \XSolidBrush & \CheckmarkBold & \CheckmarkBold & \CheckmarkBold & 50.66 & 60.73& 52.47 & 37.07\\ \midrule
(V) & \CheckmarkBold & \XSolidBrush & \XSolidBrush & \XSolidBrush & 52.16 & 69.26 & 54.18 & 30.34\\
(VI) & \CheckmarkBold & \CheckmarkBold & \XSolidBrush & \CheckmarkBold & 52.76 & 70.32 & 55.17 & 29.97 \\
(VII) & \CheckmarkBold & \CheckmarkBold & \CheckmarkBold & \XSolidBrush & 52.83 & 69.34 &55.58 &30.82\\
(VIII) & \CheckmarkBold & \CheckmarkBold & \CheckmarkBold & \CheckmarkBold &52.86 & 68.65 & 55.61 & 31.67 \\




\bottomrule
\end{tabular}
\vspace{-2mm}
\end{table}


\subsection{Comparison with State-of-the-art Methods}
We compare our method against existing algorithms, including MiSLAS~\cite{zhong2021improving}, LADE~\cite{hong2021disentangling}, ACE~\cite{cai2021ace}, DRO-LT~\cite{samuel2021distributional}, PaCo~\cite{cui2021parametric}, DiVE~\cite{he2021distilling}, IB+Focal~\cite{park2021influence}, VS~\cite{kini2021label}, TCM~\cite{xu2021towards}, DisAlign~\cite{zhang2021distribution}, and GistNet~\cite{liu2021gistnet} on datasets mentioned in Table~\ref{tab:data}.

\noindent\textbf{Results on CIFAR10-LT/100-LT.}
Table~\ref{tab:comp_cifar} summarizes the details, showing that all existing cutting-edge long-tailed approaches produce promising results. In comparison to previous approaches, our compound batch normalization properly accommodates the distribution shift between training and testing, resulting in a significant improvement. In particular, we attain an average precision of 85.0\%/53.3\%, whereas the available best of the rest is 82.8\%/52.0\% under imbalance condition $\rho=100$ for CIFAR10-LT and CIFAR100-LT, respectively. 
The compound normalization method can be considered of as a novel genre which is orthogonal to existing re-sampling, re-weighting strategies. More analysis can be noticed in Table~\ref{tab:comp-rsrw}.

\begin{table}
\centering
\caption{Comparison against other methods on CIFAR10-LT and CIFAR100-LT. ResNet32 is used as the backbone model.}\label{tab:comp_cifar}

\resizebox{\linewidth}{!}{

\begin{tabular}{l | c c c| c c c } 
\toprule
\multirow{2}{*}{\textbf{Methods}} &  \multicolumn{3}{c|}{\textbf{CIFAR10-LT}} & \multicolumn{3}{c}{\textbf{CIFAR100-LT}} \\ \cmidrule(l){2-4} \cmidrule(l){5-7 }

& $\rho=100$ & $\rho=50$ & $\rho=10$  & $\rho=100$ & $\rho=50$ & $\rho=10$ \\

\midrule
MiSLAS~\cite{zhong2021improving}  &
82.1 & 85.7 & 90.0 & 47.0 & 52.3 & 63.2 \\
LADE~\cite{hong2021disentangling}  & - & - & -&45.4 & 50.5 & 61.7 \\
ACE~\cite{cai2021ace} & 81.4 &84.9 &- & 49.6 & 51.9 &- \\
DRO-LT~\cite{samuel2021distributional} & - & -& -&47.3 & 57.6 & 63.4 \\
PaCo~\cite{cui2021parametric}   & - & -& -& 52.0 & 56.0 & 64.2 \\
DiVE~\cite{he2021distilling}   &- & -& -& 45.4 & 51.1 & 62.0 \\
SSD~\cite{li2021self} &- & -&- & 46.0 & 50.5 & 62.3 \\
IB+Focal~\cite{park2021influence}   & 78.0 & 82.4 & 87.9 & 45.0 & 48.9 & 59.5 \\
VS~\cite{kini2021label}   & 80.8 &- & -&43.5 & -& -\\
TCM~\cite{xu2021towards}   & 82.8 & 84.3 & 89.7 & 45.5 & 51.1 & 61.3 \\
\textbf{Ours}   & \textbf{85.0} & \textbf{88.7} & \textbf{91.8} & \textbf{53.3} & \textbf{60.0} & \textbf{65.4}  \\
\bottomrule
\end{tabular}
}
\end{table}

\noindent\textbf{Results on ImageNet-LT.}
On ImageNet-LT, Table~\ref{tab:comp_imagenet} presents detailed experimental results for comparisons with contemporary state-of-the-art algorithms using the ResNet50. We observe that PaCo~\cite{cui2021parametric} achieves comparable results (57.0\%) that are slightly inferior to ours (57.4\%). However, PaCo extends the contrastive framework MoCo~\cite{he2020momentum,chen2020improved} by introducing a new momentum encoder, which is much more demanding than our approach, to alleviate the long-tail problem. In addition to PaCo, our approach outperforms the remaining methods with an remarkable margin.

\begin{table}
\centering
\caption{Comparison against other methods on ImageNet-LT. ResNet50 is used as the backbone model.}\label{tab:comp_imagenet}

\setlength\tabcolsep{2mm}
\begin{tabular}{l | c} 
\toprule
Methods & Top-1 Accuracy \\ \midrule
MiSLAS~\cite{zhong2021improving} & 52.7 \\
DisAlign~\cite{zhang2021distribution} & 52.9 \\
LADE~\cite{hong2021disentangling} & 52.0 \\
ACE~\cite{cai2021ace} & 54.7 \\
DRO-LT~\cite{samuel2021distributional}  & 53.5 \\
PaCo~\cite{cui2021parametric} & 57.0 \\
TCM~\cite{xu2021towards} & 48.4 \\
Ours & 57.4\\
\bottomrule
\end{tabular} 
\end{table}

\noindent\textbf{Results on Places-LT.}
Places-LT is a long-tail variation of Places2~\cite{zhong2021improving}. The studies are carried out using the backbone ResNet152 which is initialized with network parameters pre-trained on ImageNet. Table~\ref{tab:comp-abide} demonstrates that our method consistently surpasses the state-of-the-art results with notable gains. 

\begin{table}
\centering
\caption{Comparison against other methods on Places-LT. ResNet152 is used as the backbone model.}\label{tab:comp-abide}

\setlength\tabcolsep{2mm}
\begin{tabular}{l | c} 
\toprule
Methods & Top-1 Accuracy \\ \midrule
MiSLAS~\cite{zhong2021improving} & 40.4 \\
DisAlign~\cite{zhang2021distribution} & 39.3 \\ 
LADE~\cite{hong2021disentangling} & 38.8 \\
PaCo~\cite{cui2021parametric} & 41.2 \\
GistNet~\cite{liu2021gistnet} & 39.6 \\
Ours & 42.7\\
\bottomrule
\end{tabular} 
\end{table}

\noindent\textbf{Results on iNaturalist2018.}
We examine our approach on the real-world long-tailed dataset iNaturalist 2018. 
Table~\ref{tab:comp_inat} shows the experimental results. Our method outperforms contemporary state-of-the-art approaches such as PaCo~\cite{cui2021parametric}, and ACE~\cite{cai2021ace}. The results indicate that our approach can handle extremely unbalanced fine-grained data in real-world scenarios despite the enormous number of classes.
\begin{table}
\centering
\caption{Comparison against other methods on iNaturalist 2018. ResNet50 is used as the backbone model.}\label{tab:comp_inat}

\setlength\tabcolsep{2mm}
\begin{tabular}{l | c} 
\toprule
Methods & Top-1 Accuracy \\ \midrule
MiSLAS~\cite{zhong2021improving} & 71.6 \\
DisAlign~\cite{zhang2021distribution} & 70.6 \\
LADE~\cite{hong2021disentangling} & 70.0 \\
ACE~\cite{cai2021ace} & 72.9 \\
DRO-LT~\cite{samuel2021distributional} & 69.7 \\
PaCo~\cite{cui2021parametric} & 73.2 \\
DiVE~\cite{he2021distilling} & 71.7 \\
SSD~\cite{li2021self} & 71.5 \\
IB+Focal~\cite{park2021influence} & 65.4 \\
GistNet~\cite{liu2021gistnet} & 70.8 \\
TCM~\cite{xu2021towards} & 69.2 \\
Ours & 74.8 \\
\bottomrule
\end{tabular} 
\vspace{--1mm}
\end{table}


\section{Conclusion}
This paper presents a compound batch normalization approach based on a mixture of Gaussian distributions that can comprehensively describe the feature space while avoiding the dominance of head classes. A moving average based expectation maximization (EM) method is introduced to capture the statistical variables for compound Gaussian distributions. The EM algorithm is sensitive to initialization and may easily get trapped in local minima. To tackle these issues, we build a dual-path learning approach that incorporates split feature normalization to diversify the Gaussian distributions. Extensive results on frequently used datasets show that the proposed method surpasses existing methods in long-tailed image classification by a considerable margin. To conclude, we present a novel perspective to address the imbalance issue at the feature level, which is inspiring to future work on this topic.

\section*{ACKNOWLEDGMENTS}
\begin{sloppypar}
This work is supported in part by the National Natural Science Foundation of China under Grant No. 62106235, 62003256, 61876140, 61976250, 62027813, U1801265, and U21B2048, in part by the Exploratory Research Project of Zhejiang Lab under Grant No. 2022PG0AN01, in part by the Zhejiang Provincial Natural Science Foundation of China under Grant No. LQ21F020003, in part by Open Research Projects of Zhejiang Lab under Grant No. 2019kD0AD01/010,  in part by the Guangdong Basic and Applied Basic Research Foundation under Grant No. 2020B1515020048, and in part by Mindspore which is a new deep learning computing framework\footnote{\url{https://www.mindspore.cn/}}.
\end{sloppypar}

\bibliographystyle{ACM-Reference-Format}
\bibliography{sample-base}

\appendix
\section{Appendix}\label{sm:algo}
\begin{algorithm}[htp]
\caption{Algorithm for the calculation process of the generalized batch normalization.} \label{alg:gbn}
\SetAlgoLined
\SetKwInput{Inputs}{Input}
\Inputs{Features: $\mathbf x\in \mathbb R^{D \times N}$; $M$ sets of Gaussian statistic variables: $\{\tau_j\}_{j=1}^M$, $\{\bm\mu_j\}_{j=1}^M$, and $\{\bm\Sigma_j\}_{j=1}^M$. }
\SetKwInput{Output}{Output}
\Output{Normalized features; updated Gaussian statistic variables.}
\begin{algorithmic}[1]

\STATE Standardize $\mathbf x$ with Gaussian distributions: 
$\hat{\textbf x}^j_i=\bm\Sigma^{-\frac{1}{2}}_j(\mathbf x_i-\bm\mu_j)$, for $i \in[1,N]$ and $j \in [1,M]$;

\STATE Calculate probabilities of input features in Gaussian distributions:
$w_{ij}= \frac{ \tau_j f(\mathbf x_i,\bm\mu_j,\bm\Sigma_j) }{\sum_{k=1}^M \tau_k f(\mathbf x_i,\mu_k,\Sigma_k) }$; 

\STATE Aggregate results of normalization branches: $\mathbf x^\prime_i = \sum_{j=1}^M w_{ij}(\gamma_j \hat{\textbf{x}}^{(j)}_i + \beta_j)$;

\IF{\textit{training}}
    \STATE Calculate temporary Gaussian statistic variables: $\hat{\tau}_j^c = \frac{1}{N} \sum_{i=1}^N w_{ij}$, $\hat{\bm \mu}_j^c = \frac{\sum_{i=1}^N w_{ij} \mathbf x_i}{\sum_{i=1}^N w_{ij}}$, and $\hat{\bm \sigma}_j^c = \frac{\sum_{i=1}^N w_{ij} (\mathbf x_i-\hat{\bm \mu}_j^c)^2}{\sum_{i=1}^N w_{ij}}$;
    \STATE Update permanent Gaussian statistic variables: 
    $\tau_j \gets \lambda^c \tau_j + (1-\lambda^c) \hat{\tau}_j^c$,
    $\bm\mu_j \gets \lambda^c\bm\mu_j+(1-\lambda^c)\hat{\bm \mu}_j^c$,
    and $\bm\Sigma_j \gets \lambda^c \bm\Sigma_j+(1-\lambda^c)\textrm{diag}(\hat{\bm \sigma}_j^c)$.
\ENDIF
\end{algorithmic}
\end{algorithm}
\begin{algorithm}[t]
\caption{Algorithm for the calculation process of the split batch normalization.} \label{alg:sbn}
\SetAlgoLined
\SetKwInput{Inputs}{Input}
\Inputs{Features: $\mathbf x\in \mathbb R^{D \times N}$, and their labels: $\mathbf y \in \mathbb R^N$; $M$ sets of class labels: $\{\mathbb C^j\}_{j=1}^M$; $M$ sets of Gaussian statistic variables: $\{\tau_j\}_{j=1}^M$, $\{\bm\mu_j\}_{j=1}^M$, and $\{\bm\Sigma_j\}_{j=1}^M$. }
\SetKwInput{Output}{Output}
\Output{Normalized features; updated Gaussian statistic variables.}
\begin{algorithmic}[1]

\STATE Initialize feature groups: $\mathbf x^j=[]$, for $j\in[1,M]$;

\FOR{$i=1$ \KwTo $N$}
    \STATE Retrieve the index of label class set $s_i$ which $\mathbf x_i$'s label belongs to;
    \STATE Append $\mathbf x_i$ into $\mathbf x^{s_i}$;
\ENDFOR

\FOR{$j=1$ \KwTo $M$}
    \STATE  Normalize $\mathbf x^j$ with the $j$-th Gaussian distribution: 
$\hat{\textbf x}^j=\bm\Sigma^{-\frac{1}{2}}_j(\mathbf x^j-\bm\mu_j)$;
    \STATE Redistribute $\hat{\mathbf x}^j$ with the affine parameters of the $j$-th normalization branch: $\mathbf x^{j\prime} = \gamma_j \hat{\textbf{x}}^j + \beta_j$;
    \IF{\textit{training}}
        \STATE  Calculate temporary Gaussian statistic variables: $\hat{\bm \mu}_j^s = \sum_{i=1}^{N_j} \mathbf x_i^j / N_j$, and $\hat{\bm \sigma}_j^s = \sum_{i=1}^{N_j} (\mathbf x_i^j-\hat{\bm \mu}_j^s)^2/N_j$, where $N_j$ denotes the number of features in  $\mathbf x^j$;
        \STATE Update permanent Gaussian statistic variables: 
    $\bm\mu_j \gets \lambda^s\bm\mu_j+(1-\lambda^s)\hat{\bm \mu}_j^s$,
    and $\bm\Sigma_j \gets \lambda^s \bm\Sigma_j+(1-\lambda^s)\textrm{diag}(\hat{\bm \sigma}_j^s)$.
    \ENDIF
\ENDFOR
\end{algorithmic}
\end{algorithm}









\end{document}